\documentclass[letterpaper]{article} 
\usepackage[preprint]{Iris}  
\usepackage[hyphens]{url}  
\usepackage{graphicx} 
\usepackage{natbib}  
\usepackage{caption} 
\usepackage{algorithm}
\usepackage{algorithmic}

\usepackage{amsmath}
\usepackage{amssymb}
\usepackage{booktabs}
\usepackage{multirow}
\usepackage{placeins}

\title{Beyond Solution-Centric Search: Adaptive Inquiry and Knowledge Revision for Autonomous ML Engineering}
\author{
    Shaokang Fu\textsuperscript{\rm 1},
    Yulong Tao\textsuperscript{\rm 1},
    Linbo Jin\textsuperscript{\rm 1}\corresponding,
    Jiarong Zhao\textsuperscript{\rm 2},\\
    Qiming Shi\textsuperscript{\rm 3},
    Tianjun Pan\textsuperscript{\rm 4},
    Haonan Li\textsuperscript{\rm 1},
    Chengyu Wang\textsuperscript{\rm 1},\\
    Jia Wu\textsuperscript{\rm 5},
    Chengfu Huo\textsuperscript{\rm 1}
}
\affiliations{
    \textsuperscript{\rm 1}Alibaba Group\quad
    \textsuperscript{\rm 2}East China Normal University\quad
    \textsuperscript{\rm 3}Zhejiang University\\
    \textsuperscript{\rm 4}Fudan University\quad
    \textsuperscript{\rm 5}Independent Researcher
}

\begin{document}

\maketitle

\begin{abstract}
Long-horizon autonomous research tasks such as machine learning engineering
require systems to make interdependent decisions under a limited budget.
Existing LLM-based agents typically organize candidate-solution improvement
through tree, graph, or chain structures, meaning that the search process
determines how information is acquired and managed. We call this design solution-centric
search and propose instead the information paradigm, in which an evolving
information state represents the system's understanding of the task and guides
solution improvement.
We instantiate this paradigm in \textbf{Iris}, an inquiry--revision loop. For
information acquisition, Iris generates local action plans from the current
information state and uses epistemic actions to probe decision-critical
unknowns without modifying the retained solution. For information management,
Iris synthesizes observations across experiments into task knowledge composed
of revisable claims with explicit scope and status. It updates this knowledge
as new evidence arrives and constructs each decision context from raw evidence,
structured summaries, or task knowledge at the required level of detail.
On MLE-Bench, Iris attains a $64.9\%$ any-medal rate under a
12-hour budget, the highest among compared systems. Across four tasks spanning
harness engineering and model post-training, Iris also demonstrates
cross-domain generalization.
\end{abstract}

\section{Introduction}

Machine learning engineering (MLE) is a demanding instance of long-horizon
autonomous research. It requires interdependent decisions about data
preparation, method design, implementation, and evaluation, all under a
fixed budget \cite{gottweis2025aicoscientist,huang2024mlagentbench,hong2025datainterpreter,li2025autokaggle}. Recent LLM-based agents
address this challenge through different search and refinement protocols,
including tree search
\cite{jiang2025aide,zhu2026mlmaster}, graph search
\cite{toledo2025aira}, targeted component refinement
\cite{nam2025mlestar}, and budget-aware multi-branch search
\cite{chen2026mars}. Despite differences in implementation, these methods
organize progress around candidate solutions, with their protocols determining
how candidates are generated, modified, and compared. We refer to this shared
design as the \emph{solution-centric search paradigm}. In this paradigm,
candidate-solution search remains the organizing principle, and information
handling is designed to support it.

\begin{figure}[!t]
    \centering
    \includegraphics[width=\columnwidth]{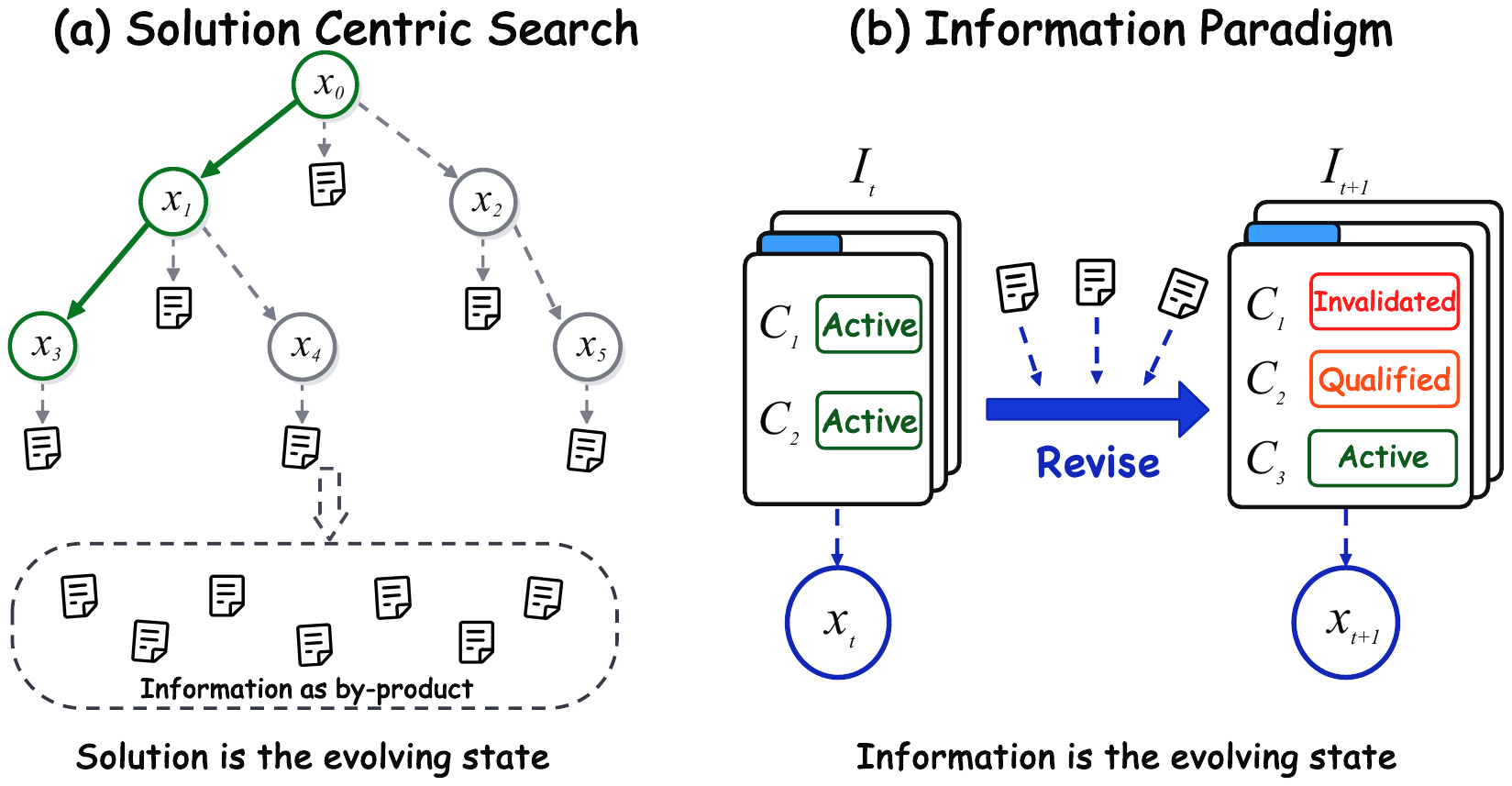}
    \caption{Two paradigms for long-horizon autonomous research. (a) The
    solution-centric search paradigm treats candidate solutions as the
    evolving state, while information produced during candidate generation and
    evaluation accumulates as a by-product. (b) The information paradigm treats
    \(\mathcal{I}_t\) as the evolving state. Evidence revises the claims it
    contains, and the revised state guides updates to the retained solution
    \(x_t\).}
    \label{fig:motivation}
\end{figure}

However, organizing research around candidate solutions leaves the system's
evolving understanding in a secondary role. In long-horizon research, early
conclusions may be
overturned by later evidence, direction selection relies on understanding
accumulated across experiments \cite{mitchener2025kosmos,tang2025chemagent,wei2025evomemory,suzgun2026dynamic}, and a limited
budget makes blind trial-and-error increasingly costly. The information
available to the system is therefore central to its subsequent decisions. We
formalize this view as the \emph{information paradigm}, which centers research
on the \emph{information state}, a continually revised representation of what
the system has learned about the task. Changes in this state guide subsequent
solution improvement.

From this perspective, existing methods exhibit two limitations, as illustrated
in Figure~\ref{fig:motivation}.
\textbf{Information acquisition is constrained.} Predefined tree, graph,
or chain structures constrain how the next action is selected from the current
research state. Existing actions still serve to modify and evaluate candidate
solutions and thus provide limited support for probing a decision-critical unknown
without producing another candidate.

\textbf{Information management remains shallow.} Existing memory mechanisms
preserve experiment records or validated successes for later use \cite{ji2026infinimemory}. Yet
cross-experiment findings are not explicitly maintained as revisable task-level
claims with recorded provenance, scope, and status. This limits conflict
resolution, evidence synthesis, and access to information at different levels
of detail.

We present \textbf{Iris} (\textbf{I}nquiry and
\textbf{R}evision over the \textbf{I}nformation \textbf{S}tate), which
organizes the research process as a loop of inquiry and
revision over the information state. For information acquisition, Iris
generates local action plans from the current information state, allowing the
action topology to adapt as research proceeds. It also distinguishes
interventional actions from \emph{epistemic actions}, which gather
discriminative evidence without modifying the retained solution $x_t$. For
information management, Iris synthesizes observations from multiple experiments
into revisable task knowledge with explicit scope and status. At each decision
point, it provides raw evidence, structured summaries, or task knowledge
according to the information required.

On MLE-Bench \cite{chan2025mlebench}, Iris attains a $64.9\%$
any-medal rate under a 12-hour budget, the highest among the compared systems.
Across four tasks spanning harness engineering and model post-training, Iris
also demonstrates cross-domain generalization.

Our contributions are threefold:
\begin{itemize}
    \item We characterize existing methods as a solution-centric search
    paradigm and propose the information paradigm, which organizes
    research around the evolution of the information state and reveals
    limitations in information acquisition and management.
    \item We instantiate the paradigm in \textbf{Iris}, an inquiry--revision
    loop with two components. Information acquisition combines adaptive action
    topology with epistemic actions, while information management combines
    cross-experiment knowledge revision with multi-granularity access.
    \item We evaluate Iris on MLE-Bench and four cross-domain
    tasks spanning harness engineering and model post-training. Under a 12-hour
    budget, Iris achieves the highest overall any-medal rate among the compared
    systems on MLE-Bench; results across the four additional tasks demonstrate
    cross-domain generalization.
\end{itemize}

\section{Related Work}
\subsection{LLM-Based MLE Agents}
General autonomous-science systems span research ideation, experiment
execution, analysis, and reporting \cite{lu2024aiscientist,schmidgall2025agentlab,yamada2025aiscientistv2}. Within MLE,
LLM-based agents optimize executable solutions through different search
and refinement protocols. AIDE, AIRA, and MLE-STAR employ tree search, graph
search, and targeted component refinement, respectively, while R\&D-Agent introduces
researcher--developer collaboration
\cite{jiang2025aide,toledo2025aira,nam2025mlestar,yang2025rdagent}.
Recent systems extend this paradigm with budget-aware search and memory
mechanisms that accumulate experiment records or retain validated successes
\cite{chen2026mars,zhu2026mlmaster,zhang2026gome}. In these methods,
candidate-solution exploration follows the chosen search structure or protocol.
Iris instead treats information acquisition and management as explicit design
objectives.

\subsection{Adaptive Information Acquisition}
Under this solution-centered organization, information arises mainly as a
by-product of modifying and evaluating a candidate. In related settings, agents
explicitly seek evidence through diagnostic reasoning, hypothesis testing, and
conjecture falsification
\cite{yang2024sweagent,boiko2023coscientist,huang2025popper}.
Active learning and Bayesian experimental design further treat acquisition as
a decision problem
\cite{settles2009activelearning,chaloner1995bayesian}, but typically assume a
predefined query or experiment space. For open-ended MLE, Iris generates local
action topologies from the evolving information state and plans epistemic
actions that target decision-critical unknowns and gather discriminative
evidence without modifying the retained solution.

\subsection{Agent Memory and Knowledge Revision}
Reflexion and Self-Refine carry verbal feedback across iterations
\cite{shinn2023reflexion,madaan2023selfrefine}; MARS and ML-Master~2.0
accumulate experiment records at multiple scales
\cite{chen2026mars,zhu2026mlmaster}; A-MEM dynamically links and updates
memory entries, while GOME maintains a validated success memory
\cite{xu2025amem,zhang2026gome}. Beyond MLE, MemGPT manages tiered memory,
while CoALA formalizes modular memory and action spaces for language agents
\cite{packer2023memgpt,sumers2024coala}. However, these mechanisms do not
represent evidence across experiments as revisable task-level claims with
explicit provenance, scope, and status, nor do they update such claims when
subsequent evidence contradicts them. Iris maintains this structure as
revisable task knowledge and
constructs decision-specific contexts from raw evidence, structured summaries,
or task knowledge.

\section{Method}

\begin{figure*}[!t]
    \centering
    \includegraphics[width=0.98\textwidth]{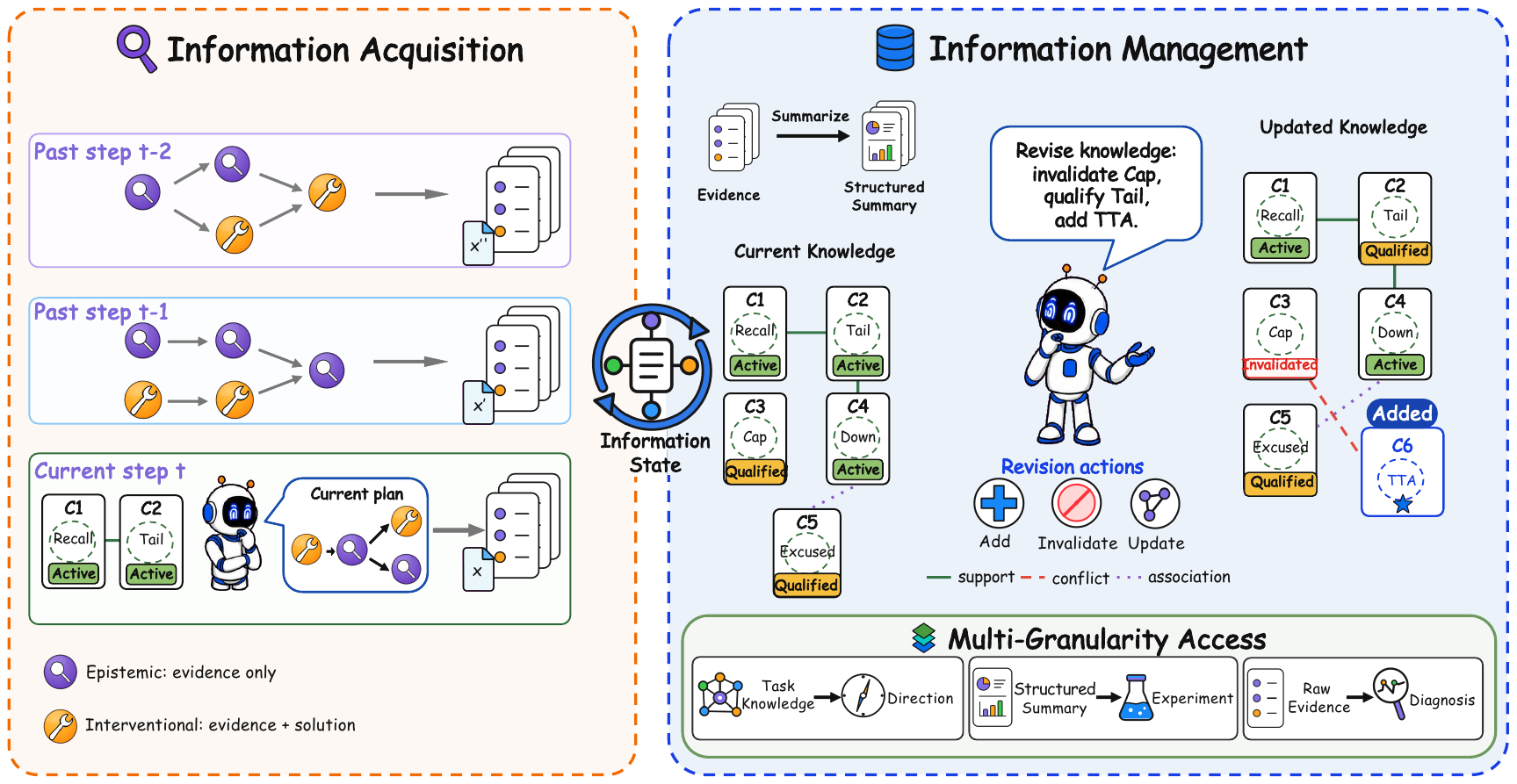}
    \caption{Overview of Iris. Left: inquiry generates local action plans that
    update \(\mathcal{I}_t\) through interventional and epistemic actions. Top
    right: Iris uses accumulated evidence to revise task knowledge. Bottom
    right: each decision receives the required information granularity. The
    revised state guides the next plan.}
    \label{fig:iris-overview}
\end{figure*}

\subsection{Problem Formulation}

Given a long-horizon MLE task \(q=(\mathcal{D}_q,J_q,B)\), where
\(\mathcal{D}_q\) contains the task description and available data, \(J_q\)
is the task evaluation function, and \(B\) is the total research budget, the
agent repeatedly plans, executes, and evaluates research actions within the
budget. This process forms a research trajectory \(\tau\) and produces a final
solution \(x_{\mathrm{out}}(\tau)\). The objective is
\begin{equation}
    \max_{\tau}\;
    J_q\!\left(x_{\mathrm{out}}(\tau)\right)
    \quad
    \text{s.t.}\quad
    T(\tau)\leq B.
    \label{eq:objective}
\end{equation}
Here, \(T(\tau)\) denotes the total time required to execute trajectory
\(\tau\). The evaluation function \(J_q\) is a protected, held-out
criterion: it is queried only once, externally, on the final solution and is
never accessible during research. Throughout the trajectory the agent
instead relies on a fixed development evaluator \(F_q\) to compare and retain
candidate solutions. Its validation set and metric are either provided by the
task or constructed from the available data before optimization.

The research trajectory \(\tau\) comprises a sequence of interleaved decision
and execution steps. At decision step \(t\), the agent's decision state is
\begin{equation}
\begin{aligned}
    s_t
        &= (x_t,\mathcal{I}_t,b_t),\\
    \mathcal{I}_t
        &= (\mathcal{E}_t,\mathcal{S}_t,\mathcal{K}_t).
\end{aligned}
\label{eq:problem-formulation}
\end{equation}
Here, \(x_t\) is the retained executable solution, \(x_0\) is the initial
valid executable artifact available at the start of research, and
\(b_t\) is the remaining research budget. The task-local information state
\(\mathcal{I}_t\) comprises the experimental and analytical evidence
\(\mathcal{E}_t\) accumulated thus far, per-action structured summaries
\(\mathcal{S}_t\), and the task knowledge \(\mathcal{K}_t\) synthesized from
evidence across experiments.

Based on \(s_t\), the agent selects and executes subsequent research actions.
Their outcomes update the task-local information state and may change the
retained solution, while consuming research budget and yielding the next
decision state \(s_{t+1}\).
This process continues until the research budget is exhausted; the
solution retained at termination constitutes \(x_{\mathrm{out}}(\tau)\).

\subsection{Overview}

Iris organizes research as a loop of inquiry and revision. Inquiry performs
information acquisition, while revision performs information management.
The two cores operate together throughout the research process
(Figure~\ref{fig:iris-overview}). The full loop is summarized in
Algorithm~\ref{alg:iris}: at each decision step the agent reads the current
information state \(\mathcal{I}_t\) and remaining budget \(b_t\), executes a
local plan that extends the evidence \(\mathcal{E}\) and summaries
\(\mathcal{S}\), revises the task knowledge \(\mathcal{K}\), and retains a
valid solution using development feedback. The updated state conditions the
next decision, and the terminal retained solution is returned. The loop is
implemented by five independently invoked LLM roles (a Planner, an Executor, an
Analyst, a Summarizer, and a Knowledge Manager), each with a role-specific
prompt. The same inquiry--revision loop is used across domains, while each task
supplies its own data, evaluator, and executable artifact.

\begin{algorithm}[t]
\caption{The Iris inquiry--revision loop over the information state.}
\label{alg:iris}
\small
\begin{algorithmic}[1]
\REQUIRE task data $\mathcal{D}_q$, initial valid artifact $x_0$, budget $B$
\REQUIRE fixed development evaluator $F_q$
\ENSURE solution $x_{\mathrm{out}}(\tau)$
\STATE $s_0=(x_0,(\varnothing,\varnothing,\varnothing),B)$;\ $t\leftarrow 0$
\WHILE{$b_t>0$}
  \STATE $(\mathcal{A}_t,\prec_t)\leftarrow\operatorname{Plan}(\mathcal{D}_q,s_t)$ \COMMENT{Planner}
  \FOR{$a_i\in\mathcal{A}_t$ by topological layer}
    \STATE $\omega_i\leftarrow\{o_j:a_j\prec_t a_i\}$ \COMMENT{predecessor outputs}
    \STATE $o_i=(e_i,\widetilde{x}_i)\leftarrow\operatorname{Execute}(a_i;s_t,\omega_i)$
    \STATE $\sigma_i\leftarrow\operatorname{Summarize}(a_i,o_i)$ \COMMENT{Summarizer}
  \ENDFOR
  \STATE $\mathcal{E}_{t+1}\leftarrow\mathcal{E}_t\cup\{e_i:a_i\in\mathcal{A}_t\}$
  \STATE $\mathcal{S}_{t+1}\leftarrow\mathcal{S}_t\cup\{\sigma_i:a_i\in\mathcal{A}_t\}$
  \STATE $\mathcal{Z}_t\leftarrow\operatorname{SelectEvidence}(\mathcal{E}_{t+1},\mathcal{S}_{t+1},\mathcal{K}_t)$
  \STATE $\mathcal{K}_{t+1}\leftarrow\operatorname{Revise}(\mathcal{K}_t,\mathcal{Z}_t)$ \COMMENT{Knowledge Manager}
  \STATE $\mathcal{V}_t\leftarrow\operatorname{Valid}(\{x_t\}\cup\{\widetilde{x}_i:a_i\in\mathcal{A}_t^{\mathrm{int}}\})$
  \STATE $x_{t+1}\leftarrow\arg\max_{x\in\mathcal{V}_t}F_q(x)$
  \STATE $\mathcal{I}_{t+1}\leftarrow(\mathcal{E}_{t+1},\mathcal{S}_{t+1},\mathcal{K}_{t+1})$
  \STATE $s_{t+1}\leftarrow(x_{t+1},\mathcal{I}_{t+1},b_t-\Delta T_t)$;\ $t\leftarrow t+1$
\ENDWHILE
\RETURN $x_{\mathrm{out}}(\tau)\leftarrow x_t$ \COMMENT{terminal retained solution}
\end{algorithmic}
\end{algorithm}

Here, \(\operatorname{Valid}\) removes candidates that fail to execute or
to produce a task-valid output. Because the valid retained solution \(x_t\) is
included at every step, \(\mathcal{V}_t\) is nonempty.

\subsection{Information Acquisition}

Information acquisition concerns how Iris produces the task evidence needed
for later decisions. It has two components: adaptive action topology and the
partition of actions into interventional and epistemic types.

\subsubsection{Adaptive Action Topology}
Iris does not fix the chain, tree, or graph structure of the full research
trajectory in advance. At decision step \(t\), the agent generates a local
action plan conditioned on the task data \(\mathcal{D}_q\) and the current
state \(s_t\):
\begin{equation}
    \operatorname{Plan}(\mathcal{D}_q,s_t)
        = (\mathcal{A}_t,\prec_t).
    \label{eq:organization}
\end{equation}
Here, \(\mathcal{A}_t\) is the set of research actions in the current local
plan and \(\prec_t\subseteq\mathcal{A}_t\times\mathcal{A}_t\) encodes
dependencies between actions. If \(a_i\prec_t a_j\), then \(a_j\) can be
executed only after \(a_i\) completes and may use the evidence, candidate
solution, or other artifacts produced by \(a_i\). A local plan may be
sequential, parallel, or mixed. Actions in the same topological layer may
execute concurrently when resources permit. The Planner emits only actions
that can be scheduled within \(b_t\). The resulting local-plan wall-clock
duration \(\Delta T_t\) satisfies \(\Delta T_t\leq b_t\).
The topology of the full trajectory thus emerges from successively generated
local plans.

\subsubsection{Epistemic Actions}
Iris partitions the research actions in a local plan into interventional
actions and epistemic actions:
\(\mathcal{A}_t=\mathcal{A}_t^{\mathrm{int}}\cup\mathcal{A}_t^{\mathrm{epi}}\).
Interventional actions construct or modify candidate solutions and evaluate
them; epistemic actions produce discriminative evidence through analysis or
diagnosis without modifying the currently retained solution. The outcomes
of completed predecessors are
\(\omega_i=\{o_j:a_j\prec_t a_i\}\). Given these outputs, the execution
operator returns
\begin{equation}
\operatorname{Execute}(a_i;s_t,\omega_i)=
\begin{cases}
    (e_i,\varnothing),
    & a_i\in\mathcal{A}_t^{\mathrm{epi}},\\
    (e_i,\widetilde{x}_i),
    & a_i\in\mathcal{A}_t^{\mathrm{int}}.
\end{cases}
    \label{eq:action-effects}
\end{equation}
We denote either output tuple by \(o_i\), with
\(\widetilde{x}_i=\varnothing\) for an epistemic action. Here, \(e_i\) is the
evidence produced and \(\widetilde{x}_i\) the candidate solution of an
interventional action; the latter is compared with the retained solution
using only the fixed development evaluator \(F_q\). The Analyst
executes epistemic actions, whereas the Executor executes
interventional actions. Typical instances
of epistemic actions include exploratory data analysis, error
stratification, feature-interaction probing, validation reliability checks,
and computational profiling. These examples do not form a predefined taxonomy;
specific epistemic actions emerge from the agent's assessment of gaps in the
current information state.

Each epistemic action centers on a \emph{decision-critical unknown}: a question
whose possible answers would lead to qualitatively different subsequent
interventions. The agent introduces an epistemic action when the current
evidence does not distinguish among possible next steps, when progress stalls
for unclear reasons, or when an unexpected result must be traced to its source.
When the available evidence is sufficient to choose the next intervention,
research proceeds directly with interventional actions. The Planner allocates
the remaining budget across the two action types with no hard cap on either.

\subsection{Information Management}

Information acquisition adds raw evidence to \(\mathcal{E}_t\). Information
management organizes this evidence into structured summaries and revisable
task knowledge while preserving access to all three levels. It determines how
new evidence revises \(\mathcal{K}_t\) and which level of information is
supplied at each decision. After each interventional or epistemic action, the
Summarizer converts its output into
\(\sigma_i=\operatorname{Summarize}(a_i,o_i)\). The summary records the
attempt, observation, and implication and is appended to
\(\mathcal{S}_{t+1}\).

\subsubsection{Knowledge Revision}
Iris revises task knowledge by synthesizing evidence across experiments.
A single structured summary describes one research action and cannot by itself
reveal failure modes shared across experiments, reconcile conflicting
findings, or integrate complementary ones. Iris therefore represents task
knowledge as a set of interrelated claim records with explicit evidential
provenance:
\begin{equation}
\begin{aligned}
    \mathcal{K}_t
        &= (\mathcal{C}_t,\mathcal{R}_t),\\
    c_i
        &= (\mathrm{claim}_i,\mathrm{scope}_i,
        E_i^{+},E_i^{-},\mathrm{status}_i).
\end{aligned}
    \label{eq:knowledge-claim}
\end{equation}
Here, \(\mathcal{C}_t\) is the set of task-knowledge claim records and
\(\mathcal{R}_t\) captures support, conflict, and association relations
among them. Each record \(c_i\) contains a claim \(\mathrm{claim}_i\), its
scope of applicability \(\mathrm{scope}_i\), supporting-evidence pointer set
\(E_i^{+}\), refuting-evidence pointer set \(E_i^{-}\), and current status
\(\mathrm{status}_i\). The status is active when the claim remains supported
under its stated scope, qualified when later evidence narrows that scope, and
invalidated when the claim is contradicted or no longer applies.

Once the evidence and summary sets have been extended, cross-experiment
synthesis uses the summaries to locate relevant evidence, consults the
corresponding raw observations when needed, and revises claim records and their
relations accordingly:
\begin{equation}
\begin{aligned}
    \mathcal{Z}_t
        &= \operatorname{SelectEvidence}
        (\mathcal{E}_{t+1},\mathcal{S}_{t+1},\mathcal{K}_t),\\
    \mathcal{K}_{t+1}
        &= \operatorname{Revise}
        (\mathcal{K}_t,\mathcal{Z}_t).
\end{aligned}
    \label{eq:knowledge-revision}
\end{equation}
Here, \(\mathcal{Z}_t\) is the evidence selected in light of the current task
knowledge \(\mathcal{K}_t\). It includes evidence that supports or refutes
existing claims, cross-experiment results that call for joint explanation, and
evidence sufficient to form new claims. A revision may apply any combination
of three operations. Add introduces new claim records. Update modifies existing
records or their relations by adjusting claims, scopes, or evidence pointers,
linking or merging complementary claims, or recording conflicts among them.
When an update narrows a claim's scope, its status may change from active to
qualified. Invalidate retires claims contradicted or rendered inapplicable by
later evidence.

\subsubsection{Multi-Granularity Access}
For each role invocation, Iris constructs a decision-specific context from
\(\mathcal{I}_t=(\mathcal{E}_t,\mathcal{S}_t,\mathcal{K}_t)\). For direction
selection, the context primarily contains task knowledge. For experiment
design, it contains relevant structured summaries. For anomaly diagnosis, it
includes raw evidence reached through provenance links. The pointers
\(E_i^{+}\) and \(E_i^{-}\) link each claim to the raw evidence that supports
or refutes it.

\section{Experiments}

We evaluate Iris on the full MLE-Bench, four cross-domain tasks spanning
harness engineering and model post-training, controlled component and
backbone studies, and a trajectory analysis of inquiry and knowledge
revision.

\begin{table*}[!t]
\centering
\small
\setlength{\tabcolsep}{2.0pt}%
\begin{tabular}{llrccccccc}
\toprule
& & & \multicolumn{4}{c}{Any-medal rate by complexity} &
\multicolumn{3}{c}{Other metrics} \\
\cmidrule(lr){4-7}\cmidrule(lr){8-10}
Method & Backbone & Time & Low & Medium & High & All & Valid & Med+ & Gold \\
& & (h) & (\%) & (\%) & (\%) & (\%) & (\%) & (\%) & (\%) \\
\midrule
AIBuildAI & Claude-Opus-4.6 & 24 & $77.3{\pm}0.0$ & $\underline{61.4{\pm}0.9}$ & $\mathbf{46.7{\pm}0.0}$ & $\underline{63.1{\pm}0.4}$ & $\mathbf{100.0{\pm}0.0}$ & $71.1{\pm}1.2$ & $25.8{\pm}0.4$ \\
MARS+ & Gemini-3-Pro & 24 & $\underline{78.8{\pm}1.5}$ & $60.5{\pm}1.5$ & $\underline{44.4{\pm}2.2}$ & $62.7{\pm}0.8$ & $\mathbf{100.0{\pm}0.0}$ & $\underline{74.2{\pm}0.9}$ & $\underline{33.8{\pm}0.4}$ \\
ML-Master 2.0 & DeepSeek-V3.2-Speciale & 24 & $75.8{\pm}1.5$ & $50.9{\pm}3.5$ & $42.2{\pm}2.2$ & $56.4{\pm}2.5$ & $95.6{\pm}1.2$ & $63.1{\pm}1.2$ & $19.6{\pm}0.9$ \\
MARS & Gemini-3-Pro & 24 & $74.2{\pm}1.5$ & $52.6{\pm}3.0$ & $37.8{\pm}2.2$ & $56.0{\pm}1.5$ & $\underline{98.7{\pm}0.0}$ & $65.8{\pm}1.6$ & $31.1{\pm}0.4$ \\
Leeroo & Gemini-3-Pro-Preview & 24 & $68.2{\pm}2.6$ & $44.7{\pm}1.5$ & $40.0{\pm}0.0$ & $50.7{\pm}1.3$ & $50.7{\pm}1.3$ & $50.7{\pm}1.3$ & $21.3{\pm}2.0$ \\
MLE-STAR-Pro-1.5 & Gemini-2.5-Pro & 24 & $68.2{\pm}2.6$ & $34.2{\pm}1.5$ & $33.3{\pm}0.0$ & $44.0{\pm}1.3$ & $93.8{\pm}0.4$ & $52.9{\pm}1.6$ & $19.1{\pm}1.8$ \\
FamouAgent & Gemini-2.5-Pro & 24 & $62.1{\pm}1.5$ & $36.8{\pm}1.5$ & $33.3{\pm}0.0$ & $43.6{\pm}0.9$ & $96.9{\pm}1.2$ & $51.6{\pm}1.2$ & $22.7{\pm}0.8$ \\
MLE-STAR-Pro-1.0 & Gemini-2.5-Pro & 12 & $66.7{\pm}1.5$ & $25.4{\pm}0.9$ & $31.1{\pm}2.2$ & $38.7{\pm}0.8$ & $94.2{\pm}1.8$ & $52.0{\pm}2.0$ & $19.1{\pm}1.2$ \\
Gome & GPT-5 & 12 & $68.2{\pm}2.6$ & $21.1{\pm}1.5$ & $22.2{\pm}2.4$ & $35.1{\pm}0.4$ & $96.0{\pm}0.0$ & $45.3{\pm}0.0$ & $16.4{\pm}0.8$ \\
AIRA-Dojo & o3 & 24 & $55.0{\pm}1.5$ & $22.0{\pm}1.2$ & $21.7{\pm}1.1$ & $31.6{\pm}0.8$ & $97.5{\pm}0.3$ & $45.5{\pm}0.8$ & $17.3{\pm}0.4$ \\
ML-Master & DeepSeek-R1 & 12 & $48.5{\pm}1.5$ & $20.2{\pm}2.3$ & $24.4{\pm}2.2$ & $29.3{\pm}0.8$ & $93.3{\pm}1.3$ & $44.9{\pm}1.2$ & $17.3{\pm}0.8$ \\
AIDE & o1-preview & 24 & $35.9{\pm}1.9$ & $8.5{\pm}0.4$ & $11.7{\pm}1.3$ & $17.1{\pm}0.6$ & $82.8{\pm}1.1$ & $29.4{\pm}1.3$ & $9.4{\pm}0.8$ \\
\midrule
\textbf{Iris (ours)} & Claude-Opus-4.6 & $\mathbf{12}$ & $\mathbf{80.3{\pm}1.5}$ & $\mathbf{64.0{\pm}0.9}$ & $\underline{44.4{\pm}2.2}$ & $\mathbf{64.9{\pm}0.4}$ & $\mathbf{100.0{\pm}0.0}$ & $\mathbf{76.0{\pm}0.8}$ & $\mathbf{39.1{\pm}0.4}$ \\
\bottomrule
\end{tabular}
\caption{Full MLE-Bench results on 75 tasks under reported 12- or 24-hour
budgets. Iris results are mean $\pm$ SEM over three runs; best and
second-best results are in \textbf{bold} and underlined.}
\label{tab:mlebench-full}
\end{table*}

\subsection{Experimental Setup}

\paragraph{Benchmarks and metrics.}
MLE-Bench \cite{chan2025mlebench} is an end-to-end MLE benchmark of
75 Kaggle competitions across three complexity levels (Low, Medium, and
High). We evaluate the full set with a 12-hour budget per task and report
official any-medal rates by complexity and overall, together with
valid-submission (Valid), above-median (Med+), and gold-medal (Gold) rates.
Iris results are mean $\pm$ SEM over three runs.

\paragraph{Cross-domain tasks.}
Beyond MLE, we evaluate two harness-engineering and two model-post-training
tasks. On $\tau^3$-Banking
\cite{shi2026tauknowledge,barres2025tau2bench}, which requires
policy-grounded tool use, we optimize the agent harness on 20 development
tasks and report average binary reward on 77 held-out tasks. On BrowseComp
\cite{wei2025browsecomp}, which requires multi-step web search and evidence
synthesis, we optimize a minimal ReAct-style harness
\cite{yao2023react} on 50 development questions and report accuracy on 300
held-out questions.

For post-training, we evaluate Iris on two autonomous model-improvement tasks
targeting HumanEval \cite{chen2021humaneval} and GSM8K
\cite{cobbe2021gsm8k}, following the task formulation of PostTrainBench
\cite{rank2026posttrainbench}. Starting from Qwen3-1.7B, each method develops a
training recipe and submits the resulting checkpoint. During optimization, each
evaluator call is limited to 150 examples; final evaluation uses all 164
HumanEval problems (pass@$1$) and all 1,319 GSM8K test examples (exact match).

\paragraph{Baselines.}
On MLE-Bench, we compare against recent 12- and 24-hour systems, including
AIDE \cite{jiang2025aide}, AIRA-Dojo \cite{toledo2025aira}, MLE-STAR
\cite{nam2025mlestar}, MARS and MARS+ \cite{chen2026mars}, ML-Master and
ML-Master~2.0 \cite{liu2025mlmaster,zhu2026mlmaster}, Gome
\cite{zhang2026gome}, FamouAgent \cite{li2025fmagent}, AIBuildAI
\cite{zhang2026aibuildai}, and Leeroo \cite{nadafian2026kapso}. We take scores
from the cited papers and official leaderboard
\cite{chan2025mlebench}.

On cross-domain tasks, we compare with Codex and Claude Code under a matched
6-hour autonomous-optimization protocol. All methods receive identical initial
artifacts, task descriptions, evaluators, editable files, and data access.
Both baselines run in official \texttt{/goal} mode with high reasoning effort
and no manual intervention. The Initial row reports performance before
optimization: the original harness for harness-engineering tasks and the
unmodified base model for post-training tasks
\cite{rank2026posttrainbench,du2026datamaster}. On harness-engineering tasks,
all methods optimize only on development feedback, with final scores produced
on protected held-out tests.

\paragraph{Implementation details.}
For the main evaluation, all Iris roles use Claude-Opus-4.6. In the
cross-domain studies and component ablations, Iris's optimization roles
use Qwen3.7-Max, while the Codex and Claude Code optimization agents use
GPT-5.5 and Claude Opus 4.8, respectively. For both harness-engineering
tasks, the optimized harness uses Qwen3.7-Max as its fixed task executor
for all methods. Each ablation run has a 6-hour wall-clock budget, and
its final submission is scored on the official private test set. Each task is
executed on 23 vCPUs, 234\,GB of RAM, and a single NVIDIA H20-3e GPU with
144\,GB of GPU memory.

\subsection{Main Results}

\paragraph{Full MLE-Bench.}
Under a 12-hour budget, Iris attains a $64.9\%$ overall any-medal rate and a
$39.1\%$ gold-medal rate, both the highest among the compared systems
(Table~\ref{tab:mlebench-full}). It produces valid submissions on all 75 tasks,
and $76.0\%$ of its submissions exceed the human median Kaggle score. Across
difficulty levels, Iris ranks first on low- and medium-complexity tasks, with
any-medal rates of $80.3\%$ and $64.0\%$, and ties for second on high-complexity
tasks at $44.4\%$. AIBuildAI provides a direct model-matched comparison because
it also uses Claude Opus 4.6. Iris raises the overall any-medal rate from
$63.1\%$ to $64.9\%$ and the gold-medal rate from $25.8\%$ to $39.1\%$ while
reducing the research budget from 24 to 12 hours.

\begin{table}[!t]
\centering
\small
\setlength{\tabcolsep}{2.0pt}
\begin{tabular}{@{}lcccc@{}}
\toprule
& \multicolumn{2}{c}{Harness engineering} & \multicolumn{2}{c}{Post-training} \\
\cmidrule(lr){2-3}\cmidrule(lr){4-5}
Method & $\tau^3$-Banking & BrowseComp & HumanEval & GSM8K \\
\midrule
Initial & $15.8$ & $38.3$ & $9.0$ & $8.5$ \\
Codex & $20.7$ & $57.5$ & $36.3$ & $52.3$ \\
Claude Code & $26.3$ & $42.8$ & $41.5$ & $54.0$ \\
\textbf{Iris} & $\mathbf{30.3}$ & $\mathbf{67.7}$ & $\mathbf{47.0}$ & $\mathbf{68.4}$ \\
\bottomrule
\end{tabular}
\caption{Cross-domain results under matched initial artifacts and 6-hour
budgets. Entries are single-run percentages on held-out harness tests and
full post-training benchmarks; metrics are average reward, accuracy,
pass@$1$, and exact match, respectively.}
\label{tab:cross-task}
\end{table}

\paragraph{Cross-domain transfer.}
Beyond MLE, Iris outperforms both coding agents on all four tasks, improving
over the stronger baseline on each task by $4.0$ to $14.4$ percentage points
(Table~\ref{tab:cross-task}). On harness engineering, it reaches $30.3\%$
average reward on $\tau^3$-Banking and $67.7\%$ accuracy on BrowseComp. These
gains show that the inquiry--revision loop can optimize agent behavior under
policy-grounded tool use and multi-step evidence synthesis. On model
post-training, Iris reaches $47.0\%$ on HumanEval and $68.4\%$ on GSM8K,
showing that the same loop can optimize training recipes and model checkpoints.
The results extend Iris's optimization capability beyond MLE.

\subsection{Component Ablations}

The main 12-hour evaluation on all 75 tasks measures Iris's end-to-end
benchmark performance, while the ablation study examines the contribution
of each component under a controlled setting. Using Qwen3.7-Max, we evaluate
Full Iris and three ablated variants for six hours on all 15 MLE-Bench
competitions with datasets no larger than 60\,MB. This Small-Data group
contains 10 low- and 5 medium-complexity tasks. On larger-data competitions,
data processing and model training consume more of a six-hour run, leaving
too few research iterations for component effects to become observable;
ablations on the Big-Data group therefore require more than six hours per task
(Table~\ref{tab:tier}).

On each task, all variants start from the same initial artifact and use the
same model, evaluator, hardware allocation, and wall-clock budget; all
non-ablated settings are held fixed. Each configuration is run once per task,
and we report valid-submission, above-human-median, gold-medal, and any-medal
rates. We ablate three components:
(1) w/o Adaptive Topology removes the adaptive organization of
action dependencies while retaining state-conditioned action selection;
(2) w/o Epistemic Actions removes actions that probe
decision-critical unknowns without modifying the retained solution; and
(3) w/o Information Management removes persistent task knowledge, leaving
the Planner with recent raw reports but without cross-experiment knowledge
revision or multi-granularity access to prior evidence.

\begin{table}[t]
\centering
\small
\setlength{\tabcolsep}{3pt}
\begin{tabular}{@{}llccccc@{}}
\toprule
Group & Data size & Low & Med. & High & Total & Budget \\
\midrule
Small-Data & $\leq 60$\,MB & 10 & 5 & 0 & 15 & $\leq 6$\,h \\
Big-Data & $\geq 180$\,MB & 12 & 33 & 15 & 60 & $> 6$\,h \\
\bottomrule
\end{tabular}
\caption{Task groups by data size and MLE-Bench complexity. The 15-task
Small-Data group supports multiple research iterations within the 6-hour
ablation budget, whereas the 60 Big-Data tasks require longer runs for
component effects to become observable.}
\label{tab:tier}
\end{table}

\begin{table}[H]
\centering
\small
\setlength{\tabcolsep}{2.0pt}
\begin{tabular}{llcccc}
\toprule
Dimension & Variant & Valid & Med+ & Gold & Any medal \\
\midrule
\multirow{2}{*}{Acquisition}
 & w/o Adapt. Topo. & $100.0$ & $73.3$ & $20.0$ & $40.0$ \\
 & w/o Epistemic & $100.0$ & $73.3$ & $20.0$ & $60.0$ \\
\midrule
Management & w/o Management & $100.0$ & $80.0$ & $13.3$ & $53.3$ \\
\midrule
\multicolumn{2}{l}{\textbf{Full Iris}} & $\mathbf{100.0}$ & $\mathbf{80.0}$ & $\mathbf{20.0}$ & $\mathbf{66.7}$ \\
\bottomrule
\end{tabular}
\caption{Component ablations on all 15 Small-Data tasks under a 6-hour
budget. Entries are percentages from one run per task; Valid denotes a valid
submission, Med+ denotes a score above the human median, and Gold and Any
medal follow the official MLE-Bench thresholds. Full Iris is the unablated
reference.}
\label{tab:ablation}
\end{table}

As shown in Table~\ref{tab:ablation}, all variants maintain a $100.0\%$
valid-submission rate, indicating that the ablated components affect
research effectiveness rather than basic execution reliability.
Removing adaptive topology reduces the any-medal rate from $66.7\%$ to
$40.0\%$ and Med+ from $80.0\%$ to $73.3\%$, while the gold-medal rate
remains unchanged. Action selection remains conditioned on the current state,
but dependencies among locally selected actions no longer adapt to the
evolving information state, making it harder to coordinate available evidence
into competitive solutions. Removing epistemic actions lowers both Med+ and
the any-medal rate by $6.7$ percentage points, while the gold-medal rate
remains at $20.0\%$. Information is then acquired only through actions that
modify or evaluate candidate solutions, preventing Iris from directly
investigating decision-critical unknowns before choosing the next
intervention. Removing information management leaves Med+ unchanged at
$80.0\%$ but reduces the any-medal rate from $66.7\%$ to $53.3\%$ and the
gold-medal rate from $20.0\%$ to $13.3\%$. Recent raw reports can still
support reasonable solutions, but the absence of cross-experiment knowledge
revision makes sustained refinement less reliable.

Across the 15 Small-Data tasks, Full Iris completes approximately seven
experiments per task within six hours. The ablation setting thus provides
multiple rounds of experimental feedback through which the acquisition and
management components can affect subsequent decisions.

\FloatBarrier
\subsection{Further Analysis}

\begin{figure}[!t]
    \centering
    \includegraphics[width=0.94\columnwidth]{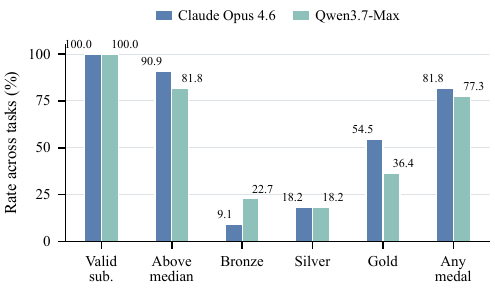}
    \caption{Backbone generality on 22 MLE-Bench Lite tasks (12 hours).
    Each backbone is evaluated with one run per task.}
    \label{fig:backbone}
\end{figure}

\paragraph{Backbone generality.}
Iris's gains could stem from its backbone rather than its system design. We
therefore keep the inquiry--revision loop, role-specific prompts, evaluator,
task set, and 12-hour budget fixed on the 22-task MLE-Bench Lite
subset, changing only the LLM used by the five roles
(Figure~\ref{fig:backbone}). Claude Opus 4.6 and Qwen3.7-Max attain any-medal
rates of $81.8\%$ and $77.3\%$, respectively, a one-task difference, and both
produce $100\%$ valid submissions. Their gold-medal rates are $54.5\%$ and
$36.4\%$, respectively, suggesting that peak solution quality is more sensitive
to model choice than overall medal attainment. This result supports our use of
Qwen3.7-Max for the controlled ablations and cross-domain studies and provides
initial evidence that the information paradigm transfers across models.

\begin{figure}[!t]
    \centering
    \includegraphics[width=0.94\columnwidth]{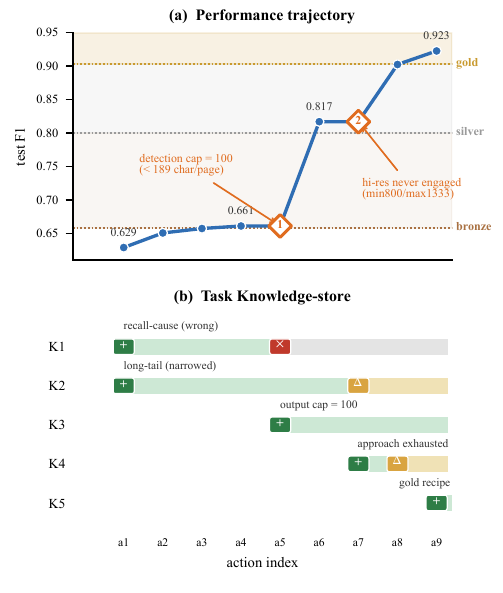}
    \caption{Inquiry--revision dynamics over actions $a_1$--$a_9$ on
    Kuzushiji recognition. \textbf{(a)} Test F1: circles mark
    interventions; diamonds mark score-neutral epistemic diagnoses that
    redirect later actions. \textbf{(b)} Aligned task-knowledge operations:
    add ($+$), update ($\Delta$), and
    invalidate ($\times$).}
    \label{fig:kuzushiji}
\end{figure}

\paragraph{Inquiry--revision dynamics.}
We examine how Iris's two cores interact within a complete optimization
trajectory on kuzushiji-recognition, a dense
detection-and-classification task scored by micro-F1 over (character,
location) pairs. The trajectory crosses all three medal thresholds along a shared
action axis (Figure~\ref{fig:kuzushiji}), allowing us to trace how
epistemic
diagnoses redirect subsequent interventions and how the same evidence
concurrently revises the task-knowledge store. All reported scores are F1
values returned by the official grader on the private test set.

From an early solution ($0.629$), routine tuning reaches a bronze plateau
($0.661$), while recall remains at $0.48$. The knowledge store attributes the
stalled performance to a modeling limitation and proposes higher resolution
and test-time augmentation as remedies (K1). The first epistemic diagnosis
instead audits the inference code and finds that a default cap of $100$
detections per image truncates dense pages. This evidence invalidates K1 and
adds the output-cap explanation as K3. Lifting the cap without retraining moves
the solution to silver ($0.817$). A second diagnosis finds that inputs are
silently downscaled, so the intended high-resolution training never takes
effect. This evidence qualifies K2 by narrowing its scope to rare classes and
updates the premature ``approach
exhausted'' record K4. Acting on the diagnosis carries the solution to gold
($0.923$), and the final configuration is added as the gold recipe K5. Both
diagnoses redirect later interventions: the first yields a gain without
retraining, and the second identifies where further training is effective.
Later decisions therefore rely on a revised understanding.

\FloatBarrier
\section{Conclusion}
We proposed the information paradigm, which attributes sustained research
improvement to changes in the information state, and instantiated it in
\textbf{Iris}. Iris organizes research as an inquiry--revision loop, acquiring
information through adaptive action topology and epistemic actions, and
managing it through cross-experiment knowledge revision and multi-granularity
access.
Within this loop, inquiry generates local plans from the current information
state and gathers evidence about decision-critical unknowns, while revision
updates task knowledge using evidence across experiments and provides the
appropriate level of information for the next planning step. On MLE-Bench,
Iris attains a $64.9\%$ any-medal rate under a 12-hour budget, the highest among
the compared systems. Across four tasks spanning harness engineering and model
post-training, Iris also demonstrates cross-domain generalization. These
results motivate treating
information acquisition and revision as central design objectives for
autonomous research systems.

\bibliography{Iris}

@inproceedings{chan2025mlebench,
  title     = {{MLE-bench}: Evaluating Machine Learning Agents on Machine Learning Engineering},
  author    = {Chan, Jun Shern and Chowdhury, Neil and Jaffe, Oliver and Aung, James and Sherburn, Dane and Mays, Evan and Starace, Giulio and Liu, Kevin and Maksin, Leon and Patwardhan, Tejal and Madry, Aleksander and Weng, Lilian},
  booktitle = {International Conference on Learning Representations},
  year      = {2025}
}

@inproceedings{yao2023react,
  title     = {{ReAct}: Synergizing Reasoning and Acting in Language Models},
  author    = {Yao, Shunyu and Zhao, Jeffrey and Yu, Dian and Du, Nan and Shafran, Izhak and Narasimhan, Karthik and Cao, Yuan},
  booktitle = {International Conference on Learning Representations},
  year      = {2023}
}

@article{lu2024aiscientist,
  title   = {The {AI} Scientist: Towards Fully Automated Open-Ended Scientific Discovery},
  author  = {Lu, Chris and Lu, Cong and Lange, Robert Tjarko and Foerster, Jakob and Clune, Jeff and Ha, David},
  journal = {arXiv preprint arXiv:2408.06292},
  year    = {2024}
}

@misc{gottweis2025aicoscientist,
  title         = {Towards an {AI} Co-Scientist},
  author        = {Gottweis, Juraj and Weng, Wei-Hung and Daryin, Alexander and Tu, Tao and Palepu, Anil and Sirkovic, Petar and Myaskovsky, Artiom and Weissenberger, Felix and Rong, Keran and Tanno, Ryutaro and others},
  year          = {2025},
  eprint        = {2502.18864},
  archivePrefix = {arXiv},
  primaryClass  = {cs.AI}
}

@misc{mitchener2025kosmos,
  title         = {{Kosmos}: An {AI} Scientist for Autonomous Discovery},
  author        = {Mitchener, Ludovico and Yiu, Angela and Chang, Benjamin and Bourdenx, Mathieu and Nadolski, Tyler and Sulovari, Arvis and Landsness, Eric C. and Barabasi, Daniel L. and Narayanan, Siddharth and Evans, Nicky and others},
  year          = {2025},
  eprint        = {2511.02824},
  archivePrefix = {arXiv},
  primaryClass  = {cs.AI}
}

@article{packer2023memgpt,
  title   = {{MemGPT}: Towards {LLM}s as Operating Systems},
  author  = {Packer, Charles and Wooders, Sarah and Lin, Kevin and Fang, Vivian and Patil, Shishir G. and Stoica, Ion and Gonzalez, Joseph E.},
  journal = {arXiv preprint arXiv:2310.08560},
  year    = {2023}
}

@article{sumers2024coala,
  title   = {Cognitive Architectures for Language Agents},
  author  = {Sumers, Theodore R. and Yao, Shunyu and Narasimhan, Karthik and Griffiths, Thomas L.},
  journal = {Transactions on Machine Learning Research},
  year    = {2024}
}

@article{jiang2025aide,
  title   = {{AIDE}: AI-Driven Exploration in the Space of Code},
  author  = {Jiang, Zhengyao and Schmidt, Dominik and Srikanth, Dhruv and Xu, Dixing and Kaplan, Ian and Jacenko, Deniss and Wu, Yuxiang},
  journal = {arXiv preprint arXiv:2502.13138},
  year    = {2025}
}

@inproceedings{toledo2025aira,
  title     = {AI Research Agents for Machine Learning: Search, Exploration, and Generalization in {MLE-bench}},
  author    = {Toledo, Edan and Hambardzumyan, Karen and Josifoski, Martin and Hazra, Rishi and Baldwin, Nicolas and Audran-Reiss, Alexis and Kuchnik, Michael and Magka, Despoina and Jiang, Minqi and Lupidi, Alisia and others},
  booktitle = {Advances in Neural Information Processing Systems},
  volume    = {38},
  year      = {2025}
}

@inproceedings{nam2025mlestar,
  title     = {{MLE-STAR}: Machine Learning Engineering Agent via Search and Targeted Refinement},
  author    = {Nam, Jaehyun and Yoon, Jinsung and Chen, Jiefeng and Shin, Jinwoo and Arik, Sercan O. and Pfister, Tomas},
  booktitle = {Advances in Neural Information Processing Systems},
  volume    = {38},
  year      = {2025}
}

@article{yang2025rdagent,
  title   = {{R\&D-Agent}: Automating Data-Driven AI Solution Building Through LLM-Powered Automated Research, Development, and Evolution},
  author  = {Yang, Xu and Yang, Xiao and Fang, Shikai and Xian, Bowen and Li, Yuante and Wang, Jian and Xu, Minrui and Pan, Haoran and Hong, Xinpeng and Liu, Weiqing and Shen, Yelong and Chen, Weizhu and Bian, Jiang},
  journal = {arXiv preprint arXiv:2505.14738},
  year    = {2025}
}

@article{chen2026mars,
  title   = {{MARS}: Modular Agent with Reflective Search for Automated AI Research},
  author  = {Chen, Jiefeng and Dalvi Mishra, Bhavana and Nam, Jaehyun and Meng, Rui and Pfister, Tomas and Yoon, Jinsung},
  journal = {arXiv preprint arXiv:2602.02660},
  year    = {2026}
}

@article{zhang2026gome,
  title   = {Reasoning as Gradient: Scaling {MLE} Agents Beyond Tree Search},
  author  = {Zhang, Yifei and Yang, Xu and Yang, Xiao and Xian, Bowen and Li, Qizheng and Fang, Shikai and Li, Jingyuan and Wang, Jian and Xu, Mingrui and Liu, Weiqing and Bian, Jiang},
  journal = {arXiv preprint arXiv:2603.01692},
  year    = {2026}
}

@article{zhu2026mlmaster,
  title   = {Toward Ultra-Long-Horizon Agentic Science: Cognitive Accumulation for Machine Learning Engineering},
  author  = {Zhu, Xinyu and Cai, Yuzhu and Liu, Zexi and Zheng, Bingyang and Wang, Cheng and Ye, Rui and Chen, Jiaao and Wang, Hanrui and Wang, Wei-Chen and Zhang, Yuzhi and Zhang, Linfeng and E, Weinan and Jin, Di and Chen, Siheng},
  journal = {arXiv preprint arXiv:2601.10402},
  year    = {2026}
}

@article{li2025fmagent,
  title   = {The {FM} Agent},
  author  = {Li, Annan and others},
  journal = {arXiv preprint arXiv:2510.26144},
  year    = {2025}
}

@article{liu2025mlmaster,
  title   = {{ML-Master}: Towards {AI}-for-{AI} via Integration of Exploration and Reasoning},
  author  = {Liu, Zexi and others},
  journal = {arXiv preprint arXiv:2506.16499},
  year    = {2025}
}

@article{zhang2026aibuildai,
  title   = {{AIBuildAI}: An {AI} Agent for Automatically Building {AI} Models},
  author  = {Zhang, Ruiyi and others},
  journal = {arXiv preprint arXiv:2604.14455},
  year    = {2026}
}

@article{nadafian2026kapso,
  title   = {{KAPSO}: A Knowledge-grounded Framework for Autonomous Program Synthesis and Optimization},
  author  = {Nadafian, Alireza and Mohammadshahi, Alireza and Yazdani, Majid},
  journal = {arXiv preprint arXiv:2601.21526},
  year    = {2026}
}

@article{huang2025popper,
  title   = {Automated Hypothesis Validation with Agentic Sequential Falsifications},
  author  = {Huang, Kexin and Jin, Ying and Li, Ryan and Li, Michael Y. and Cand\`es, Emmanuel and Leskovec, Jure},
  journal = {arXiv preprint arXiv:2502.09858},
  year    = {2025}
}

@techreport{settles2009activelearning,
  title       = {Active Learning Literature Survey},
  author      = {Settles, Burr},
  institution = {University of Wisconsin--Madison},
  number      = {1648},
  year        = {2009}
}

@article{chaloner1995bayesian,
  title   = {Bayesian Experimental Design: A Review},
  author  = {Chaloner, Kathryn and Verdinelli, Isabella},
  journal = {Statistical Science},
  volume  = {10},
  number  = {3},
  pages   = {273--304},
  year    = {1995}
}

@inproceedings{shinn2023reflexion,
  title     = {Reflexion: Language Agents with Verbal Reinforcement Learning},
  author    = {Shinn, Noah and Cassano, Federico and Gopinath, Ashwin and Narasimhan, Karthik and Yao, Shunyu},
  booktitle = {Advances in Neural Information Processing Systems},
  volume    = {36},
  year      = {2023}
}

@article{xu2025amem,
  title   = {{A-MEM}: Agentic Memory for {LLM} Agents},
  author  = {Xu, Wujiang and Liang, Zujie and Mei, Kai and Gao, Hang and Tan, Juntao and Zhang, Yongfeng},
  journal = {arXiv preprint arXiv:2502.12110},
  year    = {2025}
}

@misc{rank2026posttrainbench,
  title         = {{PostTrainBench}: Can {LLM} Agents Automate {LLM} Post-Training?},
  author        = {Rank, Ben and Bhatnagar, Hardik and Prabhu, Ameya and Eisenberg, Shira and Nguyen, Karina and Bethge, Matthias and Andriushchenko, Maksym},
  year          = {2026},
  eprint        = {2603.08640},
  archivePrefix = {arXiv},
  primaryClass  = {cs.SE}
}

@misc{du2026datamaster,
  title         = {{DataMaster}: Data-Centric Autonomous {AI} Research},
  author        = {Du, Yaxin and Yang, Xiyuan and Zhou, Zhifan and Liu, Wanxu and Lei, Zixing and Chen, Zimeng and Liu, Fenyi and Wu, Haotian and Cai, Yuzhu and Liu, Zexi and Zhu, Xinyu and Wang, WenHao and Zhang, Linfeng and Qian, Chen and Chen, Siheng},
  year          = {2026},
  eprint        = {2605.10906},
  archivePrefix = {arXiv},
  primaryClass  = {cs.LG}
}

@misc{barres2025tau2bench,
  title         = {$\tau^2$-Bench: Evaluating Conversational Agents in a Dual-Control Environment},
  author        = {Barres, Victor and Dong, Honghua and Ray, Soham and Si, Xujie and Narasimhan, Karthik},
  year          = {2025},
  eprint        = {2506.07982},
  archivePrefix = {arXiv},
  primaryClass  = {cs.AI}
}

@misc{shi2026tauknowledge,
  title         = {$\tau$-Knowledge: Evaluating Conversational Agents over Unstructured Knowledge},
  author        = {Shi, Quan and Zytek, Alexandra and Razavi, Pedram and Narasimhan, Karthik and Barres, Victor},
  year          = {2026},
  eprint        = {2603.04370},
  archivePrefix = {arXiv},
  primaryClass  = {cs.AI}
}

@misc{wei2025browsecomp,
  title         = {{BrowseComp}: A Simple Yet Challenging Benchmark for Browsing Agents},
  author        = {Wei, Jason and Sun, Zhiqing and Papay, Spencer and McKinney, Scott and Han, Jeffrey and Fulford, Isa and Chung, Hyung Won and Passos, Alex Tachard and Fedus, William and Glaese, Amelia},
  year          = {2025},
  eprint        = {2504.12516},
  archivePrefix = {arXiv},
  primaryClass  = {cs.AI}
}

@misc{cobbe2021gsm8k,
  title         = {Training Verifiers to Solve Math Word Problems},
  author        = {Cobbe, Karl and Kosaraju, Vineet and Bavarian, Mohammad and others},
  year          = {2021},
  eprint        = {2110.14168},
  archivePrefix = {arXiv},
  primaryClass  = {cs.LG}
}

@misc{chen2021humaneval,
  title         = {Evaluating Large Language Models Trained on Code},
  author        = {Chen, Mark and Tworek, Jerry and Jun, Heewoo and others},
  year          = {2021},
  eprint        = {2107.03374},
  archivePrefix = {arXiv},
  primaryClass  = {cs.LG}
}

@article{boiko2023coscientist,
  title   = {Autonomous chemical research with large language models},
  author  = {Boiko, Daniil A. and MacKnight, Robert and Kline, Ben and Gomes, Gabe},
  journal = {Nature},
  volume  = {624},
  number  = {7992},
  pages   = {570--578},
  year    = {2023}
}

@inproceedings{madaan2023selfrefine,
  title     = {Self-Refine: Iterative Refinement with Self-Feedback},
  author    = {Madaan, Aman and Tandon, Niket and Gupta, Prakhar and Hallinan, Skyler and Gao, Luyu and Wiegreffe, Sarah and Alon, Uri and Dziri, Nouha and Prabhumoye, Shrimai and Yang, Yiming and others},
  booktitle = {Advances in Neural Information Processing Systems},
  volume    = {36},
  year      = {2023}
}

@inproceedings{yang2024sweagent,
  title     = {SWE-agent: Agent-Computer Interfaces Enable Automated Software Engineering},
  author    = {Yang, John and Jimenez, Carlos E. and Wettig, Alexander and Lieret, Kilian and Yao, Shunyu and Narasimhan, Karthik and Press, Ofir},
  booktitle = {Advances in Neural Information Processing Systems},
  volume    = {37},
  year      = {2024}
}

@inproceedings{huang2024mlagentbench,
  title     = {{MLAgentBench}: Evaluating Language Agents on Machine Learning Experimentation},
  author    = {Huang, Qian and Vora, Jian and Liang, Percy and Leskovec, Jure},
  booktitle = {Proceedings of the 41st International Conference on Machine Learning},
  pages     = {20271--20309},
  year      = {2024}
}

@inproceedings{hong2025datainterpreter,
  title     = {Data Interpreter: An {LLM} Agent for Data Science},
  author    = {Hong, Sirui and Lin, Yizhang and Liu, Bang and Liu, Bangbang and Wu, Binhao and Zhang, Ceyao and Li, Danyang and Chen, Jiaqi and Zhang, Jiayi and Wang, Jinlin and Zhang, Li and Zhang, Lingyao and Yang, Min and Zhuge, Mingchen and Guo, Taicheng and Zhou, Tuo and Tao, Wei and Tang, Robert and Lu, Xiangtao and Zheng, Xiawu and Liang, Xinbing and Fei, Yaying and Cheng, Yuheng and Ni, Yongxin and Gou, Zhibin and Xu, Zongze and Luo, Yuyu and Wu, Chenglin},
  booktitle = {Findings of the Association for Computational Linguistics: ACL 2025},
  pages     = {19796--19821},
  publisher = {Association for Computational Linguistics},
  year      = {2025},
  doi       = {10.18653/v1/2025.findings-acl.1016}
}

@inproceedings{li2025autokaggle,
  title     = {{AutoKaggle}: A Multi-Agent Framework for Autonomous Data Science Competitions},
  author    = {Li, Ziming and Zang, Qianbo and Ma, David and Guo, Jiawei and Zheng, Tianyu and Liu, Minghao and Niu, Xinyao and Wang, Yue and Yang, Jian and Liu, Jiaheng and Zhong, Wanjun and Zhou, Wangchunshu and Huang, Stephen and Zhang, Ge},
  booktitle = {ICLR 2025 Workshop on Deep Learning for Code},
  year      = {2025}
}

@inproceedings{schmidgall2025agentlab,
  title     = {Agent Laboratory: Using {LLM} Agents as Research Assistants},
  author    = {Schmidgall, Samuel and Su, Yusheng and Wang, Ze and Sun, Ximeng and Wu, Jialian and Yu, Xiaodong and Liu, Jiang and Moor, Michael and Liu, Zicheng and Barsoum, Emad},
  booktitle = {Findings of the Association for Computational Linguistics: EMNLP 2025},
  pages     = {5977--6043},
  publisher = {Association for Computational Linguistics},
  year      = {2025}
}

@article{yamada2025aiscientistv2,
  title   = {The {AI} Scientist-v2: Workshop-Level Automated Scientific Discovery via Agentic Tree Search},
  author  = {Yamada, Yutaro and Lange, Robert Tjarko and Lu, Cong and Hu, Shengran and Lu, Chris and Foerster, Jakob and Clune, Jeff and Ha, David},
  journal = {arXiv preprint arXiv:2504.08066},
  year    = {2025}
}

@inproceedings{tang2025chemagent,
  title     = {{ChemAgent}: Self-updating Memories in Large Language Models Improves Chemical Reasoning},
  author    = {Tang, Xiangru and Hu, Tianyu and Ye, Muyang and Shao, Daniel and Yin, Xunjian and Ouyang, Siru and Zhou, Wangchunshu and Lu, Pan and Zhang, Zhuosheng and Zhao, Yilun and Cohan, Arman and Gerstein, Mark},
  booktitle = {International Conference on Learning Representations},
  year      = {2025}
}

@article{wei2025evomemory,
  title   = {{Evo-Memory}: Benchmarking {LLM} Agent Test-time Learning with Self-Evolving Memory},
  author  = {Wei, Tianxin and Sachdeva, Noveen and Coleman, Benjamin and He, Zhankui and Bei, Yuanchen and Ning, Xuying and Ai, Mengting and Li, Yunzhe and He, Jingrui and Chi, Ed H. and Wang, Chi and Chen, Shuo and Pereira, Fernando and Kang, Wang-Cheng and Cheng, Derek Zhiyuan},
  journal = {arXiv preprint arXiv:2511.20857},
  year    = {2025}
}

@inproceedings{suzgun2026dynamic,
  title     = {Dynamic Cheatsheet: Test-Time Learning with Adaptive Memory},
  author    = {Suzgun, Mirac and Yuksekgonul, Mert and Bianchi, Federico and Jurafsky, Dan and Zou, James},
  booktitle = {Proceedings of the 19th Conference of the European Chapter of the Association for Computational Linguistics},
  pages     = {7080--7106},
  publisher = {Association for Computational Linguistics},
  year      = {2026},
  doi       = {10.18653/v1/2026.eacl-long.333}
}

@article{ji2026infinimemory,
  title   = {Infini Memory: Maintainable Topic Documents for Long-Term {LLM} Agent Memory},
  author  = {Ji, Suozhao and Wu, Baodong and Wang, Zehao and Xia, Lei and Li, Qingping and Wang, Ruisong and Ding, Wenbo and Zhu, Zhenhua and Li, Boxun and Dai, Guohao and Wang, Yu},
  journal = {arXiv preprint arXiv:2606.10677},
  year    = {2026}
}


\end{document}